\documentclass[a4paper]{article}

\usepackage{INTERSPEECH_v2}
\usepackage[hyphens]{url}
\usepackage{hyperref}
\usepackage[linesnumbered,ruled]{algorithm2e}
\usepackage{subfig}
\title{Crowdsourcing Universal Part-Of-Speech Tags for Code-Switching}

\name{Victor Soto, Julia Hirschberg \thanks{Thanks to Morgan Ulinksi and Thamar Solorio for their useful advice and
support.  This research was funded in part by NSF CNS 1205556 "CI-ADDO-NEW:
Collaborative Research: A Repository for Annotating Multilingual Code Switched
Data" and in part by a Google Faculty Research Award.}}
\address{Columbia University, NY, USA}
\email{vsoto@cs.columbia.edu, julia@cs.columbia.edu}

\graphicspath{ {./images/} }

\begin{document}

\maketitle
% * <victor.soto.mrtez@gmail.com> 2017-03-18T14:27:59.390Z:
%
% ^.
%%JH: uhoh can we change the abstract?
%%VS: Nope.
\begin{abstract}
Code-switching is the phenomenon by which bilingual speakers
switch between multiple languages during communication.
The importance of developing language technologies for code-switching
data is immense, given the large populations that routinely
code-switch. High-quality linguistic annotations are extremely
valuable for any NLP task, and performance is often
limited by the amount of high-quality labeled data. However,
little such data exists for code-switching. In this paper, we describe
crowd-sourcing universal part-of-speech tags for the Miami
Bangor Corpus of Spanish-English code-switched speech.
We split the annotation task into three subtasks: one in which a
subset of tokens are labeled automatically, one in which questions
are specifically designed to disambiguate a subset of high frequency
words, and a more general cascaded approach for the
remaining data in which questions are displayed to the worker
following a decision tree structure. Each subtask is extended
and adapted for a multilingual setting and the universal tagset.
The quality of the annotation process is measured using hidden
check questions annotated with gold labels. The overall agreement
between gold standard labels and the majority vote is between
0.95 and 0.96 for just three labels and the average recall
across part-of-speech tags is between 0.87 and 0.99, depending
on the task.
\end{abstract}

\section{Introduction \& Previous Work}
Linguistic Code-Switching (CS) occurs when a multilingual speaker switches languages during written or spoken communication. CS typically involves two or more languages or varieties of a language and it is classified as inter-sentential when it occurs between utterances or intra-sentential when it occurs within the utterance. For example a Spanish-English speaker might say ``El teacher me
dijo que Juanito is very good at math.'' (The teacher told me that Juanito is very good at math). 

CS can be observed in various linguistic levels of representation for different language pairs: phonological, morphological, lexical, syntactic, semantic, and discourse/pragmatic switching. However, very little code-switching annotated data exists for language id, part-of-speech tags, or other syntactic, morphological, or discourse phenomena from which researchers can train statistical models. In this paper, we present an annotation scheme for obtaining part-of-speech (POS) tags for code-switching using a combination of expert knowledge and crowdsourcing. POS tags have been proven to be valuable features for NLP tasks like parsing, information extraction and machine translation \cite{och2004smorgasbord}. They are also routinely used in language modeling for speech recognition and in the front-end component of speech synthesis for training and generation of pitch accents and phrase boundaries from text \cite{taylor1998architecture, taylor1998assigning, zen2009statistical, hirschberg1990accent, watts2011unsupervised}.

With the advent of large scale machine learning approaches, the annotation of large datasets has
become increasingly challenging and expensive. Linguistic
annotations by domain experts are key to any language understanding task, but unfortunately they
are also expensive and slow to obtain. One widely adopted solution is crowdsourcing.
In crowdsourcing naive annotators submit annotations
for the same items on crowdsourcing platforms such as Amazon Mechanical Turk (AMT) and Crowdflower (CF). These are then aggregated into a single label
using a decision rule like majority vote.
Crowdsourcing allows one to obtain  annotations quickly
 at lower cost. It also raises
some important questions about the validity and quality of the annotations,
mainly: a) are aggregated labels by non-experts as good as labels by
experts? b) what steps are necessary to ensure quality? and c) how do you
explain complex tasks to non-experts to maximize output quality?
\cite{callison2010creating}.

 In \cite{snow2008cheap} the authors crowdsourced
annotations in five different NLP tasks.
To evaluate the quality of the new annotations they measured the
agreement between the gold and crowdsourced labels. Furthermore, they showed
that training a machine learning model on the crowdsourced labels yielded a
high-performing model. \cite{callison2009fast} crowdsourced translation quality
evaluations and found that by aggregating non-expert judgments it was possible to
achieve the quality expected from experts. In \cite{hsueh2009data} crowdsourcing
was used to annotate sentiment in political snippets using multiple noisy labels.
The authors showed that eliminating noisy annotators and ambiguous examples improved the 
quality of the annotations. \cite{finin2010annotating} described a crowdsourced
approach to obtaining Named Entity labels for Twitter data from  a set of four
labels using both AMT and CF. They found that a small fraction of workers
completed most of the annotations and that those workers tended to score
highest inter-annotator agreements. \cite{jha2010corpus} proposes a two-step
disambiguation task to extract prepositional phrase attachments from noisy blog data.

%%%JH: i think i would cut this para since we're not doing anything like this
%The aggregation scheme is a key component in a crowdsourcing %task. Majority
%voting is widely used but is sensitive to noisy labels.
%In \cite{hovy2013learning} the authors proposed MACE (Multi %Annotator
%Competence Estimation), an aggregation scheme based on item-%response
%models. MACE learns to identify which annotators are %trustworthy and predict
%correct labels. Similarly in \cite{rodrigues2014sequence},
%a Conditional Random Field (CRF) is used for situations where %multiple
%annotations are available but not
%actual ground truth. The algorithm proposed there was able to %simultaneously learn
%the CRF parameters, reliability of the annotators and the %estimated ground truth.

The task of crowdsourcing POS tags is challenging insofar as
POS tagsets tend to be large and the task is intrinsically sequential. This means
that  workers need to be instructed about a large number of categories and they need to 
focus on more than the word to tag, making the task potentially longer, more difficult, and
thus, more expensive. More importantly, even though broad differences between
POS tags are not hard to grasp, more subtle differences
 tend to be critically important. An example would be deciding whether  a word like "up" is  being used as a preposition ("He lives up the street") or a particle ("He lived up to the expectations.")
%particle of a phrasal verb is an adverb or an adposition. 
%%%JH: i would just say "whether a word is a preposition or a particle".  most people who read this will not be linguists and will not be familiar with the universal part-of-speech tagsets.  i'll change now.

Previous research has
tackled the task of crowdsourcing POS tags. The authors in
\cite{hovy2014experiments} collected five judgments per word in a task which
consists of reading a short context where the word to be tagged occurs, and
selecting the POS tag from a drop-down menu. Using MACE \cite{hovy2013learning} they obtained 82.6\%
accuracy and 83.7\% when restricting the number of words to be tagged using
dictionaries. In his M.S. thesis, Mainzer \cite{mainzer2011labeling} proposed
an interactive approach to crowdsourcing POS tags, where workers are assisted
through a sequence of questions to help disambiguate the tags with minimal
knowledge of linguistics. Workers following this approach for the Penn Treebank (PTB) Tagset
\cite{santorini1990part} achieved 90\% accuracy. 

In this work, we propose to 
adapt the monolingual annotation scheme from
\cite{mainzer2011labeling} to crowdsource Universal POS tags in a code-switching setting
for the Miami Bangor Corpus. Our main contributions are the following: finding mappings to the universal POS tagset, extending a monolingual annotation scheme to a code-switching setting, creating resources for the second language of the pair (Spanish) and creating a paradigm that others can adopt to annotate other code-switched language pairs.

\section{The Miami Bangor Corpus}

The Miami Bangor corpus is a conversational speech corpus recorded from
bilingual Spanish-English speakers living in Miami, FL. It includes 56
files of conversational speech from 84 speakers. The corpus consists of
242,475 words (transcribed) and 35 hours of recorded conversation. 63\%
of transcribed words are English, 34\% Spanish, and 3\% are undetermined.
The manual transcripts include beginning and end times of utterances and
per word language identification.

The original Bangor Miami corpus was automatically glossed and tagged with
POS tags using the Bangor Autoglosser \cite{donnelly2011bangor,donnelly2011using}.
The autoglosser finds the closest English-language gloss
for each token in the corpus and assigns the tag or group of tags most common
for that word in the annotated language. These tags have two main problems:
they are unsupervised and the tagset used is uncommon and not specifically
designed for multilingual text. To overcome these problems we decided to a)
obtain new part-of-speech tags through in-lab annotation and crowdsourcing and
b) to use the Universal Part-of-Speech Tagset \cite{petrov2011universal}.
%All words are glossed with the closest English-language equivalent (in
%lower case) and, where appropriate, information about parts of
%speech. English equivalents of proper names are used where they
%exist (for example, ‘Nueva_York@s:spa’ is glossed as ‘New_York’). If
%there is no English-language equivalent to a name, it is glossed
%‘name’
The Universal POS tagset is ideal for annotating code-switching corpora because
it was designed with the goal of being appropriate to any language. Furthermore, it
is useful for crowdsourced annotations because it is much smaller than other
widely-used tagsets. Comparing it to the PTB POS tagset
\cite{santorini1990part,marcus1993building}, which has a total of 45 tags,
the Universal POS tagset
%JH i thought the penn set had 52
%VS actually it has 45! 46 was incorrect
has only 17: Adjective, Adposition, Adverb, Auxiliary
Verb, Coordinating and Subordinating Conjunction,
Determiner, Interjection, Noun, Numeral, Proper Noun,
Pronoun, Particles, Punctuation, Symbol,
Verb and Other. A detailed description of the tagset can be found in
\url{http://universaldependencies.org/u/pos/}.

\section{Annotation Scheme}
The annotation scheme we have developed consists of multiple tasks: each token
is assigned to a tagging task depending on word identity, its
language and whether it is present in one of three disjoint wordlists.
The process combines a) manual annotation by computational
linguists, b) automatic annotation based on knowledge distilled from the Penn
TreeBank guidelines and the Universal Tagset guidelines, and c) and d) two
language-specific crowdsourcing tasks, one for English and one for Spanish. The pseudocode of the annotation scheme is shown in
Algorithm \ref{alg:annotation}.
\begin{algorithm}
\SetKwInOut{Input}{Input}
\underline{function RetrieveGoldUniversalTag} $(token, lang, tag)$\;
\Input{A word $token$, lang ID $lang$ and POS tag $tag$}

\uIf{IsInUniqueLists$(token, lang)$}
{
   return RetrieveAutomaticTag$(token, lang)$\;
}
\uElseIf{IsInManualAnnotationList$(token, lang)$}
{
   return RetrieveManualTag$(token, lang)$\;
}
\uElse
{
$utag$ = Map2Universal$(token, lang, tag)$\;
\uIf{IsInTSQList$(token, lang)$}
{
   utags = TokenSpecificQuestionTask$(token, 2)$\;
}
\uElse
{
   utags = QuestionTreeTask$(token, lang, 2)$\;
}
   return MajorityVote$([utag,utags])$\;
}
\caption{Pseudocode of the annotation scheme.}
\label{alg:annotation}
\end{algorithm}
Table \ref{tab:annotation_task} shows the number and percentage of tokens
tagged in each annotation task (second and third column) and the percentage of tokens that was annotated by experts in-lab, either because it was the manual task or because there was a tie in the crowdsourced task.
%Note that only 2.78\% of the corpus was
% annotated manually, by inspection; the vast majority of labels came from the crowd %block and the rest came from breaking ties from the crowdsourced
%tokens. 
%%JH: it looks to me like the largest category was the automatic tagging and in a way this was done  "in the lab" altho automatically.  i'd omit any generalization here and just go straight to the explanations.
%%JH: this  table  and these distinctions are confusing.  
%% VS: Second column is number of tokens in each category, third column is the percentage of the corpus that each category tackled, and fourth column is the % of token that ended up begin expert-annotated: in the case of the manual section, in full, but for the crowdsourced sections, because there was a 1-1-1 tie.
%%%JH: that's not what was confusing it's the distinctions being made.  hopefully that's clear now below tho.
In the next subsections we explain in detail
each one of the annotation blocks.
\begin{table}
\centering
\begin{tabular}{cccc}
\hline	
	Task & \# Tokens & \% Corpus & \% by Experts \\
\hline
	Automatic & 156845 & 56.58 & 0.00 \\
	Manual & 4,032 & 1.45  & 1.45 \\
	TSQ & 57,248 & 20.65  & 0.93 \\
	English QT & 42,545 & 15.34 & 0.32\\
	Spanish QT & 16,587 & 5.98 & 0.08 \\
\hline
	Total & 277,257 & 100  & 2.78\\
\hline
\end{tabular}
\caption{Breakdown of amount of corpus annotated per task.}
\label{tab:annotation_task}
\end{table}
All the wordlists and sets of questions and answers mentioned but not included
in the following sections are available in \url{www.cs.columbia.edu/~vsoto/cspos_supplemental.pdf}
%VS: or give a link to my website?
%%JH: good idea
%%VS: done, will upload if we submit.

\subsection{Automatically tagged tokens}
\label{sec:automatic_tokens}
For English, the PTB Annotation guidelines \cite{santorini1990part}
instructs annotators to tag a certain subset of words with a given POS tag.
We follow those instructions by mapping the fixed PTB tag to a Universal tag.
Moreover we expand this wordlist with a) English words that we found were always
tagged with the same Universal tag in the Universal Dependencies Dataset and
b) low-frequency words that we found only occur with a unique tag in the Bangor
Corpus. 

Similarly, for Spanish, we automatically tagged all
the words tagged with a unique tag throughout the Universal
Dependencies Dataset (e.g. conjunctions like `aunque', `e', `o', `y', etc.; adpositions like `a', `con', `de', etc.; and some adverbs, pronouns and numerals) and low frequency words that only occurred with one tag
throughout the Bangor corpus (e.g. `aquella', `tanta', `bastantes', etc.).
%JH: give a few examples
%VS: Done.

Given the abundance of exclamations and interjections in conversational speech,
we collected a list of frequent interjections in the corpus and tagged
them automatically as INTJ. For example: `ah', `aha', `argh', `duh',  `oh', `shh'.
%% VS: included examples.
Finally, tokens labeled as Named Entities or Proper Nouns in the original Miami
Bangor Corpus were automatically tagged as PROPN.
%%%JH: i killed some paragraphs in case we need to save space.  you can restore if we don't
\subsection{Manually tagged tokens}
\label{sec:manual_tokens}

We identified a set of English and Spanish words that we found to be particularly challenging for naive workers
to tag and which occurred in the dataset in such low frequency that we were able to have them tagged in the lab by computational linguists. Note that a question specific to each one of these tokens could have been designed for crowdsourced annotations the way it was done for the words in section \ref{sec:question_specific}. The majority of these are tokens that needed to be disambiguated between adposition and adverb in English (e.g.`above', `across', `below', `between') and between determinant and pronoun in Spanish (e.g. `algunos/as', `cu\'antos/as', `muchos/as'). 
%JH: you need some examples
%VS: Done.

\subsection{Crowdsourcing Universal Tags}
We used crowdsourcing to obtain new gold labels for every word not
manually or automatically labeled. We started with the two basic approaches discussed in
\cite{mainzer2011labeling} for disambiguating POS tags using crowdsourcing which we modified for a multilingual corpus.
In the first task 
%(subsection \ref{sec:question_specific}), 
a question and a
set of answers were designed to disambiguate the POS tag of a specific token.
In the second task 
%(subsection \ref{sec:question_tree}), 
we defined two
Question Trees (one for English and one for Spanish) that sequentially ask
non-technical questions of the workers until the POS tag is disambiguated.
These questions were designed so that the worker needs minimal knowledge of
linguistics. All the knowledge needed, including definitions, is given as
instructions or as examples in every set of questions and answers.
Most of the answers contain examples illustrating the potential uses for the
token in that answer.

Two judgments were collected from the pertinent crowdsourced task and a third
one was computed from applying a mapping from the Bangor tagset to the Universal
tagset 
%(subsection \ref{sec:mapping}). 
The new gold standard was computed as the
majority tag between the three POS tags.

\subsubsection{Token-specific questions (TSQ)}
\label{sec:question_specific}

In this task, we designed a question and multiple answers specifically
for particular word tokens. The worker was then asked to choose the answer that is the most
true in his/her opinion.
%%JH: you need an example, like this one
Below is the question we asked workers for the token `can' (Note that  users cannot
see the POS tags when they select one of the answers):
\noindent\rule[0.5ex]{\linewidth}{0.5pt}
In the context of the sentence, is `can' a verb that takes the meaning of `being able to' or `know'?
\begin{itemize}
	\item Yes. For example: `I can speak Spanish.' ({\bf AUX})
	\item No, it refers to a cylindrical container. For example: `Pass me a can of beer.' ({\bf NOUN})
\end{itemize}  
\noindent\rule[0.5ex]{\linewidth}{0.5pt}

We began with the initial list of English words and the questions developed in
\cite{mainzer2011labeling} for English. However, we added additional token-specific
questions for words that a) we thought would  be especially challenging to label
(e.g. `as', `off', `on') and b) appear frequently throughout the corpus (e.g. `anything', `something', `nothing').

We designed specific questions for a subset of Spanish words. Just as for
English, we chose a subset of most frequent words that we thought would be
especially challenging for annotation by workers like tokens that can be either adverbs or adpositions (e.g.`como', `cuando', `donde') or determiners and pronouns (e.g. `ese/a', `este/a', `la/lo')
%%JH: again, an example or 2 would be better
%%VS: Done!
We modified many of the questions proposed in \cite{mainzer2011labeling}, to
adapt them to a code-switching setting and to the universal POS tagset. 
For example, the token `no' can be an Adverb and Interjection in Spanish,
and also a Determiner in English.
Also, some of our questions required workers to choose the most accurate translations for a token in a given context:
\noindent\rule[0.5ex]{\linewidth}{0.5pt}
\noindent In the context of the sentence, would `la' be translated in English as `her' or `the'?
\begin{itemize}
	\item The (`La ni\~{n}a est\'a corriendo' becomes `The girl is running') ({\bf DET})
	\item Her (`La dije que parase' becomes `I told her to stop') ({\bf PRON})
\end{itemize}
\noindent\rule[0.5ex]{\linewidth}{0.5pt}

\subsubsection{Annotations Following a Question Tree}
\label{sec:question_tree}
In this task the worker is presented with a sequence of questions that follows 
a tree structure. Each answer selected by the user leads to the next question
until a leaf node is reached, when the token is assigned a POS tag.
We followed the basic tree structure proposed in \cite{mainzer2011labeling},
but needed to modify the trees  considerably due again to the multilingual context. For example, the new Question Tree starts by first
asking whether the token is an interjection or a proper noun. This
is very important since any verb, adjective, adverb or noun can effectively be
part of or itself be an interjection or proper noun. If the worker responds negatively,
then they are asked to follow the rest of the tree.
The resulting tree is slightly simpler than the one in \cite{mainzer2011labeling}.
This is mainly because we moved the Particle-Adverb-Adposition disambiguation
from this task into the Token-Specific Questions task. On the other hand, we
added question nodes designed to disambiguate between main verbs and auxiliary verbs.
The following is an example of the annotation task following
the English Question Tree:
\noindent\rule[0.5ex]{\linewidth}{0.5pt}
\noindent Read the sentence carefully:
\noindent ``Sabes porque I plan to move in August but I need to find a really good job.''
\noindent In the context of the sentence, is the word `good':
\begin{itemize}
	\item A Proper Noun or part of a Proper Noun.
	\item A single word used as an exclamation that expresses acknowledgement or an emotional reaction.
	\item None of the above. \checkmark
\end{itemize}
In the context, `good' is a:
\begin{itemize}
\item Noun, because it names a thing, an animal, a place, events or ideas.
\item Adjective, because it says something about the quality, quantity or the kind of noun or pronoun it refers to. \checkmark
\item Verb, because it is used to demonstrate an action or state of being.
\item Adverb, because it tells the how, where, when, when or the degree at which something is done.
\end{itemize}
Could `good' be a noun or a verb?
\begin{itemize}
\item It could be a Noun. For example, fun can be a noun as in ... or an adjective as in... 
\item It could be a Verb. For example, surprised can be a verb as in ... or an adjective as in ...
\item No, it's definitely an Adjective. \checkmark
\end{itemize}
\noindent\rule[0.5ex]{\linewidth}{0.5pt}

%\begin{figure}[h]
%\includegraphics[scale=0.4]{english_tree_example}
%\centering
%\caption{An example of annotating a word using the English Question Tree.}
%\label{fig:english_tree}
%\end{figure}

For the Spanish portion of the corpus, we modified the English subtasks still further, adapting them according to the syntactic
properties of Spanish. One of the key differences from the English tree concerns verbs in their infinitival form.
Users that choose to tag a token as verb are then asked to confirm that the
infinitival form is not a noun, and if it is not, to decide whether a verb is
acting as main verb or as an auxiliary verb (as a compound verb or periphrasis).
%Figure \ref{fig:spanish_tree} shows an example of an annotation task using the
%Spanish Question Tree.

%\begin{figure}[h]
%\includegraphics[scale=0.4]{spanish_tree_example}
%\centering
%\caption{An example of annotating a word using the Spanish Question Tree.}
%\label{fig:spanish_tree}
%\end{figure}

\subsubsection{Mapping Stage}
\label{sec:mapping}

We use the pre-annotated tag from the Bangor corpus as the third tag to 
aggregate using majority voting. To obtain it, we first cleaned the corpus
of ambiguous tags, and then defined a mapping from the Bangor tagset to
the Universal tagset. This mapping process was first published in
\cite{alghamdi2016part}.

\section{Results}

We assigned two judgments per token for each of our tasks. Before
they were allowed to begin the tasks, workers were pre-screened using a quiz of
ten questions. If two or more questions were missed during the initial quiz, the
worker was denied access to the task. Furthermore, workers were required to be
certified for the Spanish language requirement in Crowdflower. Only workers from
U.K., U.S.A., Spain, Mexico and Argentina were allowed access to the task. The tasks for the workers were designed
to present 9 questions per page plus one test question used to assess  
workers' performance. When a worker reached an accuracy lower than 85\% on these test questions, all their
submitted judgments were discarded and the task made subsequently unavailable. Every
set of 9+1 judgments was paid 5 cents (USD) for the Token-Specific Questions task and 6 cents for the Question Tree tasks. 

Table \ref{tab:agreement} shows the number of test questions for each task
and of evaluation metrics to estimate the accuracy of the annotations obtained
from the crowdsourcing workers. Taking into account all the judgments submitted
for test questions, the majority voting tag had an accuracy of 0.97-0.98 depending
on the task. These estimations are not expected to match the true accuracy we would
get from the two judgments we obtained for the rest of non-test tokens, so we
re-estimate the accuracy of the majority vote tag for every subset of one, two,
three and four judgments collected, adding the initial Bangor tag.
In this case we get an average accuracy
ranging from 0.89-0.92 with just one token to 0.95-0.96 when using four tags.
The best accuracy estimates for our POS tags are for the option of two crowdsourced
tags and the Bangor tag, for which we obtained accuracies of 0.92 to 0.94.
When looking at non-aggregated tags, the average accuracy per token
of single judgments (SJ) were observed to be between 0.87 and 0.88. Measuring the
agreement between single judgments and the majority vote (MV) per token, the average
agreement value is between 0.87 and 0.89. 
\begin{table}[h]
\centering
\begin{tabular}{|l|c|c|c|}
\hline
Task & TSQ &  Eng QT & Spa QT \\
\hline
\# Tokens & 57.2K & 42.5K & 16.6K \\
\# Test Questions & 271 & 381 & 261 \\
Avg. \# Judgments per TQ & 55.72 & 28.60 & 16.28 \\
\hline
Accuracy & 0.98 & 0.98 & 0.97 \\
Avg. Acc of SJ per TQ & 0.88 & 0.89 & 0.87\\
Avg. Agrmnt of SJ wrt MV & 0.89 & 0.90 & 0.87\\
\hline
Accuracy(1+1) &  0.89 & 0.92 &  0.91 \\
Accuracy(2+1) &  0.94 & 0.92 &  0.92 \\
Accuracy(3+1) &  0.94 & 0.96 &  0.96 \\
Accuracy(4+1) &  0.96 & 0.95 &  0.96 \\	
\hline
\end{tabular}
\caption{Accuracy and Agreement measurements per task.}
\label{tab:agreement}
\end{table}

We examine the vote split for every non-test token to obtain a measure of
confidence for the tags. We see that we consistently obtained full-confidence
crowdsourced tags on at least 60\% of the tokens for each of the tasks, reaching
70\% for the Spanish Question Tree task. The option for which one of the crowdsourced
tags was different from the other two (marked as 2-1 Bangor) on the table occurred
between 18\% and 23\% of the time depending on the task, whereas the split where the
Bangor tag was different from the crowdsourced tags (marked as 2-1 CF) occurred only
between 10.63 and 12.15\% of the time. Finally the vote was split in three different
categories only between 1.29\% and 4.51\% of the time. In those instances, the tie
was broken by in-lab annotators.
\begin{table}[h]
\centering
\begin{tabular}{|l|c|c|c|}
\hline
Task & TSQ &  English QT & Spanish QT \\
\hline
3-0 & 60.12 & 67.20 & 70.09 \\
2-1 (Bangor) & 23.20 & 19.74 & 17.98 \\
2-1 (CF) & 12.16 & 10.97 & 10.63 \\ 
1-1-1 &	4.51 & 2.09 & 1.29 \\
\hline
\end{tabular}
\caption{Voting split per task.}
\label{tab:vote_split}
\end{table}
To further evaluate the performance of the annotation process by different
tag categories, we examine the recall on the gold test questions. The recall across all tags and tasks is higher than 0.93 except for
Interjections and Adjectives for the Spanish Question Tree and Adverbs for 
the English Question Tree. Looking at the failed test questions for Adverbs,
it becomes apparent that workers had difficulty with adverbs of place that can
also function as nouns, like: `home', `west', `south', etc. For example `home' in
`right when I got home' was tagged 24 times as a Noun, and only 5 as an Adverb.
\begin{table}[h]
\centering
\begin{tabular}{|l|c|c|c|}
\hline
Task & TSQ &  Eng QT & Spa QT \\
\hline
ADV & 0.98 & 0.2 & 1.0 \\
ADJ & 1.0 & 0.97 & 0.86\\ 
ADP & 1.0 & X & X\\
AUX & 1.0 & 0.98 & 1.0 \\
CONJ & 1.0 & X & X\\
DET & 1.0 & X & X\\
INTJ & 1.0 & 1.0 & 0.78\\
NOUN & 1.0 & 1.0 & 0.96 \\
NUM & 1.0 & X & X\\
PART & 1.0 & X & X\\
PRON & 0.93 & X & X\\
PROPN & X & 1.0 & X \\	 
SCONJ & 0.96 & X & X\\
VERB & 1.0 & 0.99 & 1.0\\
\hline
Average & 0.99 & 0.88 & 0.93 \\
\hline
\end{tabular}
\caption{POS tags recall per task.}
\label{tab:recall_per_tag}
\end{table}
\section{Conclusions}
We have presented a new scheme for crowdsourcing Universal POS tagging of Spanish-English code-switched data derived from a monolingual process which also used a different tagset.  Our scheme consists of four
different tasks (one automatic, one manual, and two crowdsourced).
Each word in the corpus is sent to only one task based upon curated
wordlists. For the crowdsourced tokens, we have demonstrated that, taking
the majority vote of one unsupervised tag and two crowdsourced judgments, we obtain
highly accurate predictions. We have also shown high agreement on the predictions:
between 95 and 99\% of the tokens received two or more
votes for the same tag. Looking at the performance of each POS tag,
our predictions averaged between 0.88 and 0.93 recall depending on the task.
\clearpage
\clearpage
\newpage

\bibliographystyle{IEEEtran}
\bibliography{mybib}

\end{document}